\documentclass[letterpaper, 10 pt, conference]{ieeeconf}  % Comment this line out if you need a4paper

\IEEEoverridecommandlockouts                              % This command is only needed if 
\usepackage{graphics} % for pdf, bitmapped graphics files
\usepackage{epsfig} % for postscript graphics files
\usepackage{mathptmx} % assumes new font selection scheme installed
\usepackage{times} % assumes new font selection scheme installed
\usepackage{amsmath} % assumes amsmath package installed
\usepackage{amssymb}  % assumes amsmath package installed

\usepackage{todonotes}
\usepackage{subcaption}
\usepackage{algorithm}
\usepackage{algpseudocode}
\usepackage[hidelinks]{hyperref}
\usepackage{cleveref}
\usepackage{siunitx}

\title{\LARGE \bf
A Unified and Modular Model Predictive Control Framework for Soft Continuum Manipulators under Internal and External Constraints}

\author{Filippo A. Spinelli$^{1}$, and Robert K. Katzschmann$^{1}$% <-this % stops a space
\thanks{$^{1}$F. A. Spinelli and R. K. Katzschmann are with the Soft Robotics Lab, ETH Zurich, Tannenstrasse 3, 8092 Zürich, Switzerland
        {\tt\small \href{mailto:fspinelli@ethz.ch}{fspinelli@ethz.ch}}, 
        {\tt\small \href{mailto:rkk@ethz.ch}{rkk@ethz.ch}}}%
}

\begin{document}

\maketitle
\thispagestyle{empty}
\pagestyle{empty}

%%%%%%%%%%%%%%%%%%%%%%%%%%%%%%%%%%%%%%%%%%%%%%%%%%%%%%%%%%%%%%%%%%%%%%%%%%%%%%%%
\begin{abstract}
    
    Fluidically actuated soft robots have promising capabilities such as inherent compliance and user safety. The control of soft robots needs to properly handle nonlinear actuation dynamics, motion constraints, workspace limitations, and variable shape stiffness, so having a unique algorithm for all these issues would be extremely beneficial.
    In this work, we adapt \emph{Model Predictive Control} (\emph{MPC}), popular for rigid robots, to a soft robotic arm called SoPrA. We address the challenges that current control methods are facing, by proposing a framework that handles these in a modular manner.
    While previous work focused on Joint-Space formulations, we show through simulation and experimental results that \emph{Task-Space MPC} can be successfully implemented for dynamic soft robotic control. We provide a way to couple the \emph{Piece-wise Constant Curvature} and \emph{Augmented Rigid Body Model} assumptions with internal and external constraints and actuation dynamics, delivering an algorithm that unites these aspects and optimizes over them.
    We believe that a \emph{MPC} implementation based on our approach could be the way to address most of model-based soft robotics control issues within a unified and modular framework, while allowing to include improvements that usually belong to other control domains such as machine learning techniques.  
\end{abstract}

\section{Introduction}\label{intro}
    In this work we have adapted a powerful model-based control method, \emph{Model Predictive Control} (\emph{MPC}), to a Soft continuum Proprioceptive Arm (SoPrA)~\cite{Toshimitsu2021SoPrA:Sensing} that can perform dynamic motions and is easily scalable. Our model is based on the Piece-Wise Constant Curvature assumption and our control on the \emph{Augmented Rigid Body Model}~\cite{Katzschmann2019DynamicModel, della2020model}. 
    
    Various model-based control methods have already been applied to soft robotic arms, such as the cascaded curvature controller based on Inverse Kinematics~\cite{Katzschmann2015AutonomousManipulator}, the Curvature Dynamic Control~\cite{DellaSantina2018DynamicEnvironment} and the Sliding Mode Control (SMC)~\cite{kazemipour2021robust}. All those methods have shown promising results, but they suffer from a few issues. One is that a soft robotic arm is strongly limited by mechanical constraints both in actuation and in end-effector space; current controllers cannot deal with those limits, but \emph{MPC} is inherently suited for constraint optimization. Della Santina et al.~\cite{DellaSantina2019DynamicObstacles} have already made steps towards addressing this problem, developing a new algorithm to integrate internal constraints into the PCC control. With our work, we have provided an environment that would make use of the information extracted from an adaptation of such method to control soft arms with more control authority and awareness. 
    % Furthermore, at the cost of an increased computational complexity it is possible to include much more information. 
    
    \begin{figure}[tb]
      \centering
      \includegraphics[trim = 20mm 55mm 50mm 20mm, clip, width = \linewidth]{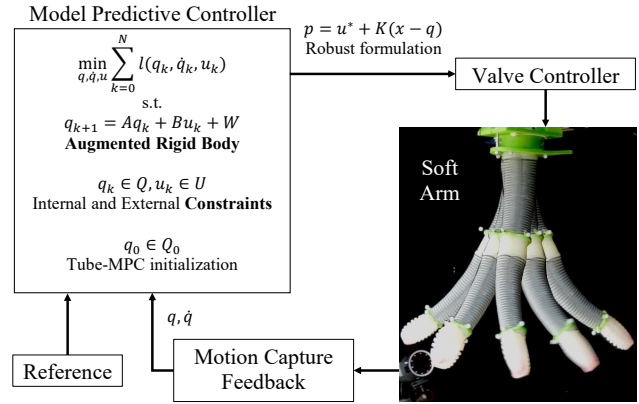}
      \caption{Our proposed controller for fluidically actuated soft arms: a robust \emph{Tube-based MPC}~\cite{Lopez2019DynamicSystems} with dynamics based on the \emph{Augmented Rigid Body Model}~\cite{Katzschmann2019DynamicModel}, internal constraints due to actuation limits and external constraints due to task-space limitations.}
      \label{fig:control_loop}
      \vspace{-10pt}
    \end{figure}
    
    Another problem of most of pneumatically actuated soft-continuous arms is the need for limiting the number of PCC sections to avoid falling into an underactuated domain. However, increasing them would lead to better performances since the kinematic and dynamic approximation would be more accurate. Also, being able to deal with underactuated systems would allow to scale up current designs and map them to models with more PCC sections. As shown for example in~\cite{Powell2015ModelWalking, falanga2018pampc}, \emph{Model Predictive Control} is usually the choice adopted for such control problems. 
    
    \emph{MPC} is an advanced model-based control technique. It is a form of optimal control with integrated feedback that directly takes into account system constraints, well suited for many model-based control problems since provides additional capabilities, including future trajectory prediction. The receding horizon approach allows to solve optimally feedback problems in a forward way, but the requirement of a finite horizon optimization leads to stability and feasibility problems, especially when dealing with disturbances. These issues have been addressed in different ways, for example with the \emph{Tube-MPC}~\cite{Lopez2019DynamicSystems} or the \emph{Indirect Feedback}~\cite{hewing2019recursively}.  
    
    \emph{MPC} has been already applied successfully to soft-robotics platforms, for example in~\cite{Best2016ModelRobots, Hyatt2020RobustMPC}. They have modeled pressure dynamics and included it directly in the optimization problem, delivering an algorithm that exploits the additional information of an augmented state-space. Best et al. have also provided comparisons between \emph{MPC} and other model-based controls~\cite{Best2021ComparingRobots}, underlying the advantages of each methods: \emph{MPC} can be as accurate as other controllers, while providing additional future knowledge and managing multiple DoFs at the same time. However, in all previous works analyzed they limited their efforts to joint-space tracking, choice that reduces possible practical applications. In this work, we moved further controlling the robot in task-space.

    Starting from the previous research results, we have implemented a robust \emph{MPC} able to deal with model uncertainties and used it for trajectory following tasks on SoPrA~\cite{Toshimitsu2021SoPrA:Sensing}. We have optimized over internal and external constraints, adopting few corrections to use standard \emph{MPC} algorithms with the \emph{Augmented Rigid Body} formulation.
    
    This work aims to improve the \emph{Augmented Rigid Body Model} technique~\cite{Katzschmann2019DynamicModel}, that takes into account geometric and inertial parameters while neglecting the pneumatic actuator dynamics and its inherent limits. By considering an augmented state-space vector as in~\cite{Best2016ModelRobots} and adapting an offline algorithm similar to~\cite{DellaSantina2019DynamicObstacles}, we can enrich our control with the knowledge needed to better tackle motion tracking problems, as well as pick-and-place tasks, and generally extend the control authority to more complex domains. \emph{MPC} has been used as unified framework to efficiently couple these techniques: kinematics and dynamics are derived from the \emph{Augmented Rigid Body Model}, pressure dynamics accounted for with an augmented state vector, task-space limitations rewritten as polytopic constraints over curvature variables. 
    
    This work contributes: 
    \begin{itemize}
        \item The \emph{Soft-Robust MPC controller} based on the \emph{Augmented Rigid Body Model} operating in task-space. The controller is fed with additional knowledge about actuation and task-space limitations to better fit into the soft-robotics control domain.
        \item A way to deal with internal constraints (actuation) and external constraints (obstacles) in a unified \emph{MPC} environment.
        \item First real-world physical validations of \emph{task-space MPC} on soft robotic arms.
        % \item A controller based on \emph{Augmented Rigid Body Model}, fed with additional knowledge to better fit into soft-robotics control domain.
        % \item A modular control framework applicable to a large variety of soft arms. 
    \end{itemize}

\section{Model} \label{model}
    \subsection{Mathematical Model}
\begin{figure}[tb]
            \centering
            \begin{subfigure}{.35\linewidth}
                \centering
                \includegraphics[trim = 0mm 0mm 0mm 0mm, clip, width = \linewidth]{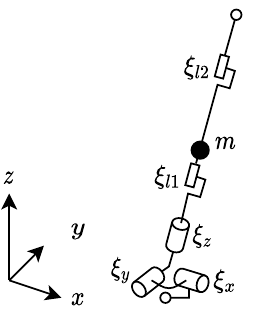}
            \end{subfigure}
            \begin{subfigure}{.35\linewidth}
                \centering
                \includegraphics[trim = 0mm 0mm 0mm 0mm, clip, width = \linewidth]{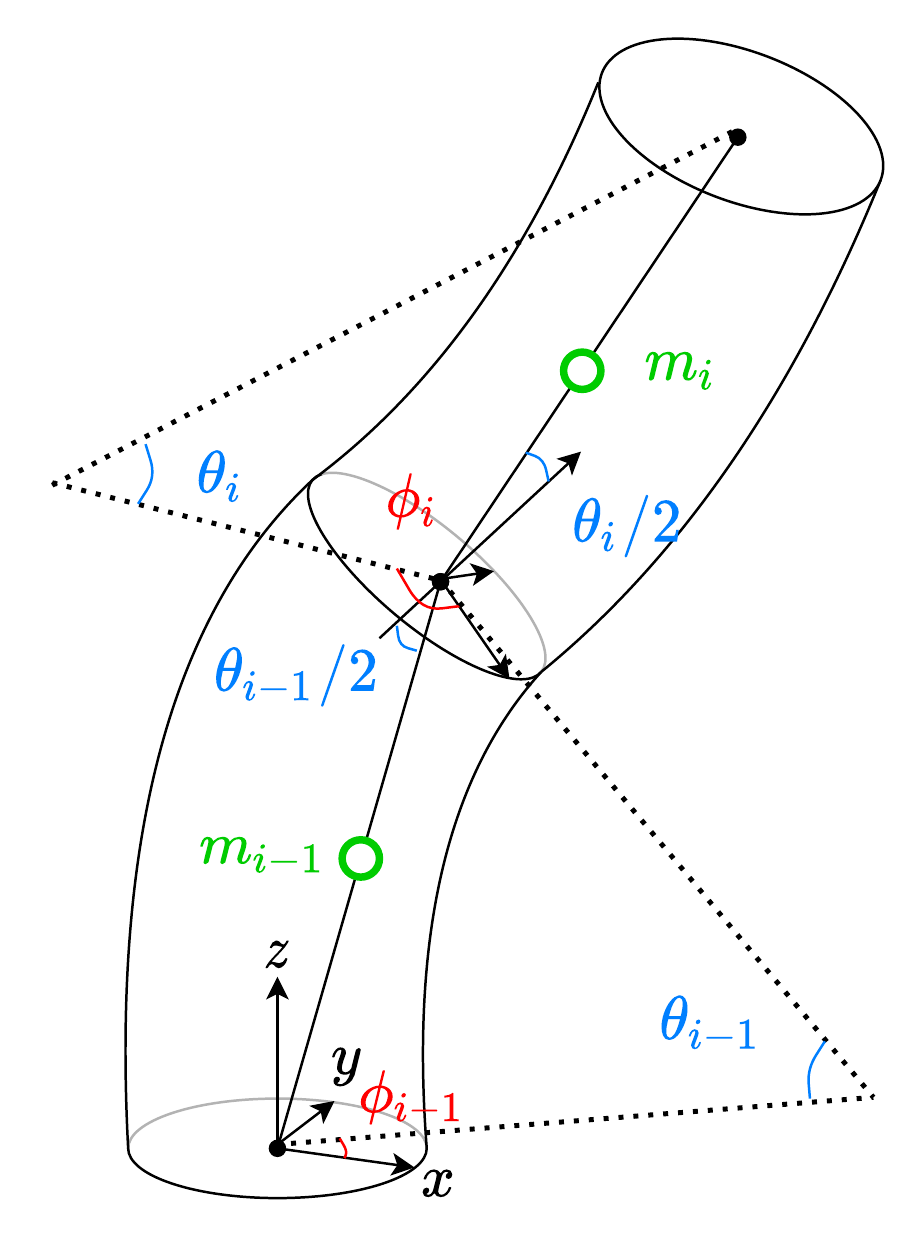}
            \end{subfigure}
            \caption{\emph{Augmented Rigid Body Model}: on the left the equivalent rigid joints and links, on the right the curvature-based space description. Adapted from~\cite{Toshimitsu2021SoPrA:Sensing}.}
            \label{figure:PCC}
            \vspace{-10pt}
        \end{figure}

        In~\cite{Toshimitsu2021SoPrA:Sensing} the kinematic and dynamic models for SoPrA have been formulated, via PCC and \emph{Augmented Rigid Body}~\cite{Katzschmann2019DynamicModel}. The dynamic equation for the soft arm is defined as:
        \begin{equation}\label{eq:dynamic_equation}
            Ap + J^Tf = B(q)\ddot{q} + c(q,\dot{q}) + g(q) + Kq + D\dot{q}
        \end{equation}
        where \emph{A} is the allocation matrix, $J^T$ the Jacobian from task-space to joint-space, \emph{B, c, g} the inertial, Coriolis and gravitational terms, \emph{K, D} additional stiffness and damping elements to account for the softness of the robot. 
        %Stiffness and damping terms have been modeled with continuum mechanics assumptions, according to the already referred paper. 
        The \emph{Augmented Rigid Body} assumption consists in an inverse mapping from a mechanical linkage made of revolute and prismatic joints in space $\xi$ to the curvature one, as in \cref{eq:ARBM_trans}. Please refer to \cite{Katzschmann2019DynamicModel, DellaSantina2018DynamicEnvironment} and \cref{figure:PCC} for more details.
        \begin{gather}\label{eq:ARBM_trans}
            B(q) = J_m^T(q)B_\xi(m(q))J_m(q) \\
            c(q, \dot{q}) = J_m^T(q)c_\xi(m(q), J_m(q)\dot{q}) \nonumber\\
            g(q) = J_m^T(q)g_\xi(m(q)) \nonumber
        \end{gather}
        Here $m(q)$ maps configuration between the two domains, and $J_m$ is the Jacobian of the mapping: 
        \begin{equation}
            J_m(q) = \frac{\partial m(q)}{\partial q}
        \end{equation}
        However, differently from~\cite{Katzschmann2019DynamicModel} only three symmetrically spaced chambers are present and the augmented state is reduced (see \cref{figure:PCC}, only 5 joints instead of 10 have been used to approximate the soft arm in 3D space). This choice leads to some adaptations in the control algorithm.
        
        According to~\cite{Toshimitsu2021SoPrA:Sensing} the robotic arm is modeled with 2 curvature parameters per section, not the bending angle $\theta$ and off-plane rotation $\phi$ but rather two derived components:
        \begin{gather}
            \theta_x = \theta cos(\phi) \quad
            \theta_y = \theta sin(\phi) 
        \end{gather}
        This approach leads to a set of variables that doesn't reach singularity in the rest straight condition. According to this definition, the curvature vector we optimize over is:
        \begin{gather}
            q = 
            \begin{bmatrix}
                \theta_{x,1} & \theta_{y,1} & \theta_{x,2} & ... & \theta_{y,N_{seg}N_{pcc}} 
            \end{bmatrix}^T \quad \text{of size}\ q_{size}
        \end{gather}
        Our arm has three pressures per segment, one per chamber, that can be controlled independently with positive values only. In order to avoid excessive constraints on actuation and the need of modifying the standard dynamic model, during control just two orthogonal and bi-directional pseudo-pressures are considered. Those are then used to geometrically reconstruct the actual input values.

    \subsection{MPC Model}
        Given \cref{eq:dynamic_equation}, in order to apply \emph{MPC} we need to get a state-space equation for system evolution. It can be achieved with the following transformation:
        \begin{gather}\label{eq:ss_matrices}
            \text{state} = 
            \begin{bmatrix}
                q \\ \dot{q}
            \end{bmatrix}, \qquad 
            \begin{bmatrix}
                \dot{q} \\ \ddot{q}
            \end{bmatrix} = 
            \bold{A} \begin{bmatrix}
                q \\ \dot{q}
            \end{bmatrix} + \bold{B}p + \bold{W}\\
            \bold{A} = 
            \begin{bmatrix}
                0_{q_{size} \times q_{size}} & I_{q_{size} \times q_{size}}\\
                -B^{-1}K & -B^{-1}D
            \end{bmatrix} \nonumber\\
            \bold{B} = 
            \begin{bmatrix}
                0_{q_{size}\times 2segments_{num} } \\
                B^{-1}A
            \end{bmatrix} \qquad
            \bold{W} =
            \begin{bmatrix}
                0_{q_{size}\times1 } \\
                -B^{-1}(c+g)
            \end{bmatrix} \nonumber
        \end{gather}
        We can write a state-space linearized system, that has to be updated each time the curvature configuration changes (so each control loop), since our formulation based on Drake~\cite{drake} would output only the numerical result of configuration-dependent matrices. One source of model errors is the Coriolis term, that is not in the common form $C(q,\dot{q})\dot{q}$ (that could be obtained for instance via Christoffel's symbols) but in a vectorial representation that can't be used to compute the \emph{A} matrix. With this strong approximation we are assuming that curvature velocity affecting the Coriolis component changes slowly along the prediction horizon, very unlikely. However, since the order of magnitude of \emph{c} term is usually much smaller than other quantities involved, the additional errors aren't extremely deteriorating for small speeds. The most critical issue in \cref{eq:ss_matrices} is that along the MPC prediction horizon the matrices won't update, so we can't consider a too long one. The inherent assumption is that state-dependent matrices would change slowly, as assumed for example also in~\cite{Hyatt2020RobustMPC}.
        
        % This problem can be solved by adding in \ref{eq:ARBM_trans}:
        % \begin{gather}
        %     C(q,\dot{q}) = J_m^T(q)B_\xi(m(q))\dot{J}_m(q,\dot{q}) + \\ J_m^T(q)C_\xi(m(q),J_m(q)\dot{q})J_m(q) \nonumber
        % \end{gather}

        Since \emph{MPC} cannot work on continuous variables due to the need to execute the optimization processes, we converted our model into discrete-time. However, the discretization would lead to additional discrepancies between the real state evolution and the predicted one. Regardless the actual sampling time chosen $T_s$, the transformations needed are: 
        \begin{gather}\label{eq:dyn_matrices}
            A_d = e^{\bold{A} \times T_s} \quad B_d = \bold{A}^{-1}(A_d - I_{2q_{size}\times 2q_{size}})\bold{B} \quad W_d = \bold{W} \times T_s 
        \end{gather}
        % Where the matrix exponential has been approximated with the fifth order expansion:
        % \begin{equation}
        %     e^{A \times T_s} = I_{2q_{size} \times 2q_{size}} + \sum_{k=1}^{5} \ \frac{1}{k!}(A \times T_s)^k
        % \end{equation}
        The whole process is computed once per control cycle. Selecting a small sampling time would result in an accurate approximation, but could produce real-time issues if the \emph{MPC} isn't able to perform all the required computations within that frame; a large sampling time would on contrary introduce large model errors and reduce the controller's noise rejection ability. 
        
        In order to design a controller able to operate in task-space, we need the forward kinematics to map the curvature variables on which the model is based to the cartesian space where motion tracking tasks are defined. Since our state-space is made by $\theta_x$ and $\theta_y$, we can just use a chain of simple rotations. The transformation we used for a two segments arm with one PCC section each is the following:
        \begin{gather}\label{eq:ee_transform}
            E_q = R_1(l_{s1}) + R_2(l_{c1}) + R_3(l_{s2}) + R_4(l_{c2}) \\
            \text{with} \quad
            R_i = R_{i-1}Rot_y(-\frac{\theta_{x,\lceil i/2 \rceil}}{2})Rot_x(-\frac{\theta_{y,\lceil i/2 \rceil}}{2}) \nonumber
        \end{gather}
        where $l_{c1,2}$ are vectors representing the lengths of the rigid connectors mounted between the segments of length $l_{s1,2}$.
        
        This kind of geometrical transformation is not considering the shrink in length that happens during bending.
        % since the respective silicon segment is almost inextensible in longitudinal direction (due to the inner backbone). 
        To take into account this problem we can use: 
        \begin{gather}
            l_{s} = 2 \ l_{0} \times \frac{sin(\frac{\theta}{2})}{\theta} \quad for \ \theta \neq 0
        \end{gather}
        However, such a computation would first need us to recover $\theta$ variable, introducing new singularities. It is therefore suggested to use only a constant scaling factor online, since approximations in the \emph{MPC} don't lead to error accumulation and wind-up, while considering the complete formula for offline computations only. 
        
        We have implemented a \emph{Robust Tube-based MPC}~\cite{Lopez2019DynamicSystems}, computing the feedback gain matrix $K$ according to the Discrete-time Algebraic Riccati Equation~\cite{kuvcera1972discrete}. The need for a robust version of MPC instead of a standard one is due to the large model mismatch and approximations, that have been considered as disturbances over the state evolution.  

        What is theoretically an offline computation, in our codebase needs to be computed online since we have different state-space matrices each control loop, when we update the state. This represents a relevant issue in terms of computation time, adding a lot of overhead. However, having selected a very conservative control rate allowed us to rarely miss the real time synchronization deadlines. 
        The second step needed for a \emph{Robust-MPC} implementation is to compute the minimum invariant set $\mathcal{E}$ under the terminal LQR control law, but our codebase would force us to do that online each iterations, something unfeasible due to the complexity of the computations. 

        We were therefore required to deal with noise bounds in an alternative way to prevent the system from being too conservative. We chose to apply the constraint tightening after the optimization, via a function that maps the output of $K(x_i-z_i)$ into a set that fulfills input constraints. But using this solution made us to lose all convergence guarantees from DARE, so in the future different approaches will be investigated.
    
    \subsection{Robust MPC formulation}
        In \cref{eq:MPC_robust} is reported our \emph{Robust MPC} scheme. We have implemented it using the CasADi C++ library~\cite{Andersson2019CasADi:Control} interfaced with IPOPT solver~\cite{ipopt2006}. This formulation is aiming for simple trajectory following tasks and has been tested successfully on various different settings.  
        \begin{gather}\label{eq:MPC_robust}
            J^*(q(k), \dot{q}(k) ) = \min_{q, \dot{q}, u} \ \mathcal{L} \\
            s.t.\quad
            \text{dynamics} \quad
            u \in \mathcal{U} \nonumber \\
            (q(N),\dot{q}(N)) \in \mathcal{X}_f \quad
            (q(0),\dot{q}(0)) \in \mathcal{X}_0 \nonumber
        \end{gather}
        The cost function in \cref{eq:MPC_cost} is defined with delta-formulation, so minimizing the distance between the reference trajectory $ref$, provided as external input parameter, and the end-effector symbolic transformation $E_q$ computed according to \cref{eq:ee_transform}. It is possible to include a multidimensional trajectory in the optimization problem to feed the \emph{MPC} with future references, so that the optimizer is able to take them into account during planning. This option is a unique advantage of \emph{MPC} over other controllers, and it is extremely useful when it is needed to track fast-varying references with abrupt variations.
        In the stage cost we have regularized both $\dot{q}$ curvature speed and the pressure input $u$, since the optimization is affected by high variance. We included also a penalization on the pressure variation, preventing the system from often choosing one of the two input limits. 
        %(happens for the Pontryagin's Minimum Principle~\cite{kirk2004optimal}). 
        Stage cost and terminal cost are both defined as quadratic functions with positive definite matrices.
        % therefore \emph{MPC} assumptions for asymptotic stability are satisfied.
        \begin{gather}\label{eq:MPC_cost}
            \mathcal{L} = (E_{q}(N)-ref(N))^TQ_N(E_{q}(N)-ref(N)) + \\ + \sum_{k=0}^{N-1} \ (E_{q}(k)-ref(k))^TQ(E_{q}(k)-ref(k)) + \nonumber \\ \dot{q}(k)^TS\dot{q}(k) + u(k)^TRu(k) + \Delta u(k)^TR\Delta u(k) \nonumber
        \end{gather}
        
        The state-space dynamic equation has been obtained according to \cref{eq:dyn_matrices}, resulting in:
        \begin{gather}\label{eq:MPC_dyn}
            \begin{bmatrix}
                q(k+1) \\ \dot{q}(k+1)
            \end{bmatrix} = 
            A_d \begin{bmatrix}
                q(k) \\ \dot{q}(k)
            \end{bmatrix} + B_d u(k) + W_d \quad \forall \ 0\leq k<N 
        \end{gather}
        For the experiments we have decided to not consider pressure dynamics to keep the computational effort light enough, however~\cite{Best2016ModelRobots} have shown as an augmented state-space helps in obtaining better results. In order to implement their dynamic constraint as \cref{eq:4states_dyn}, it is needed to define the pressure dynamic evolution, possibly empirically, and properly adapt the cost function. 
        \begin{gather}\label{eq:4states_dyn}
        \begin{bmatrix}
            \dot{q}(k+1) \\ q(k+1) \\ P_0(k+1) \\ P_1(k+1)
            \end{bmatrix} = \bar{A}_d 
            \begin{bmatrix}
                \dot{q}(k) \\ q(k) \\ P_0(k) \\ P_1(k)
            \end{bmatrix} + \bar{B}_d 
            \begin{bmatrix}
                P_{0,d}(k) \\ P_{1,d}(k)
            \end{bmatrix}
        \end{gather}
        In \cref{eq:4states_dyn} $P_{0,1}$ are the 2 chambers pressures and $P_d$ the commanded pressures.
    
        Pressure constraints in \cref{eq:MPC_pressure} model absolute limits and step variation $\Delta u$. In this way we can account for actuator internal constraints during optimization. However, since our model is linearized about the initial state, extending the constraining along the whole prediction horizon could lead to unfeasibilities. In fact we cannot capture stiffness and damping changing during the virtual state evolution since we update matrices only at the beginning of the control loop; allowing the system to produce any pressures towards the end of the horizon would limit the problems related to our static model and doesn't provide any danger since the only applied input is the first one, where limits are present. However, feasibility guarantees of the \emph{MPC} formulation are lost and our controller could fail unexpectedly during the optimization, so we needed to properly account for it.  
        \begin{gather}\label{eq:MPC_pressure}
            \mathcal{U} :\quad - \Delta u < u(0)-u_{old} < \Delta u \quad \wedge \\
            - \Delta u < u(k+1)-u(k) < \Delta u \quad \wedge \nonumber \\ \quad p_{min} < u(k) < p_{max} \quad \forall \ 0 \leq k<N/4 \nonumber
        \end{gather}
        The terminal constraint has been object of specific care, getting to use a safety filter inspired method. 
        \begin{gather}\label{eq:MPC_terminal}
            \mathcal{X}_f :\quad  \dot{q}(N) = 0
        \end{gather}
        Predictive safety filters have been introduced in reinforcement learning setting, when policy training is executed directly on the real hardware. Since the goal is to explore enough the state-space to collect valuable information, it is very likely that the system would get in dangerous situations, given a not perfectly known state evolution and unpredictable effects of new actions on unexplored domains. The idea of the safety filter is to continuously plan a trajectory able to bring back the system into a safe condition and adapt control inputs from the learning algorithm to ensure it. Refer to~\cite{JMLR:v16:garcia15a} for more information about safe RL.
        
        With SoPra we could have possible problems related to unwanted oscillations derived from the model uncertainties at high speeds. Considering a \emph{MPC} trajectory that, while minimizing the euclidean distance from the reference, ensures the existence of a control sequence to bring back the robot to a stationary condition turned out to be extremely valuable against model approximations. Particularly, this choice helps in limiting too fast pressure variations and, keeping curvature speed controlled, makes the unwanted effects due to Coriolis term discrepancy almost negligible. However, a horizon long enough has to be chosen, otherwise the robot would be too limited and would be barely able to follow any trajectories.
        
        The initial constraint is designed as required for the \emph{Robust MPC}: 
        \begin{gather}\label{eq:MPC_initial}
            \mathcal{X}_0 :\quad A_Iq(0) < Q_0 \qquad A_I\dot{q}(0) < \dot{Q}_0
        \end{gather}
        $Q_0$ and $\dot{Q}_0$ represent the offsets of a rectangular polytope centered around $q(k)$ and $\dot{q}(k)$ measured values at time $k=0$.  According to the standard \emph{Tube MPC}~\cite{Lopez2019DynamicSystems}, the control input is:
        \begin{equation}\label{tube_policy_mine}
            p^* = u(0) + K\begin{bmatrix}
                q(k) - q(0)\\ \dot{q}(k) - \dot{q}(0)
            \end{bmatrix}
        \end{equation} 
        exploiting the initialization within a neighboring set. Since we didn't tightened the constraints on pressure, being too complex online or too restrictive offline, we have then limited the amount of variation of $p^*$ from nominal input $u(0)$ at actuation phase.
        
        The optimal problem described in \cref{eq:MPC_robust} is modular: it can be adjusted with additional elements to address more complex settings. 
        As a first step, we have implemented two different ways to account for obstacles, each of them with peculiar capabilities. 
        
        The first method penalizes those states that lead to end-effector configurations too close to specified obstacles. We add the cost term shown in ~c\ref{eq:penalized_MPC} with a constant penalization $L > 0$, that smoothly decreases with distance from obstacles $o_i$. We have used an exponential function since easily differentiable during problem solving, with $l$ factor to weight distance variation. 
        However, working on-line, it cannot be used for many obstacles without increasing too much the computation time and therefore losing tracking performance. It represents a straightforward implementation to avoid unexpected sparse obstacles, but cannot be used systematically and further could affect the controller convergence. 
        \begin{gather}\label{eq:penalized_MPC}
            \mathcal{L} += \sum_{k=0}^{N-1} \ \sum_{i=0}^{N_{obs}}\frac{L}{\exp{(l(E_q(k) - o_i)^T(E_q(k) - o_i))}}
        \end{gather}
        
        The second method can account for many obstacles at the same time, since the algorithm that we derived from~\cite{DellaSantina2019DynamicObstacles} works offline. We compute a set $\mathcal{X} : \{q \ | \ A_I q < b_I\} $for the joint variables and then design a \emph{soft-MPC} based on it, as summarized in \cref{eq:soft_MPC}.
        \begin{gather}\label{eq:soft_MPC}
            \min_{q, \dot{q}, u} \ \mathcal{L} + \sum_{k=0}^{N-1} \ \epsilon(k)^TE\epsilon(k) \\
            s.t.  \quad
            \text{dynamics} \nonumber \\
            A_I q(k) < b_I + \epsilon(k) \quad \forall \ 0 \leq k < N \nonumber \\
            u \in \mathcal{U} \quad
            (q(N),\dot{q}(N)) \in \mathcal{X}_f \quad
            (q(0),\dot{q}(0)) \in \mathcal{X}_0 \nonumber
        \end{gather}
        \emph{Soft-MPC} is a \emph{MPC} formulation where some of the constraints are relaxed via the addition of a slack variable $\epsilon$, properly penalized in the cost. With this method is possible to operate under constraint configurations without the risk of incurring into unfeasibilities. However, only non-critical constraints can be softened, since failing to meet pressure requirements could lead to hardware damage. 
        
        Our implementation provides a restriction at joint-level, so multiple obstacles and further constraints can be managed. The algorithm is briefly summarized in \cref{alg:constraint_finder}. The function \emph{check\_inclusion($q_l, q_u$)} randomly tries different curvature combinations in the interval provided and checks whether obstacles are hit and targets are reached. If most of targets can be reached within a neighborhood while avoiding obstacles, then returns $True$ for the proposed interval and the previous optimal solution is updated checking the 2-norm set.   
        This approach allows to consider infinitely many constraints at each position along the arm, but can provide only pseudo-symmetric sets. Therefore, sparse obstacles cannot be handled efficiently, while it is really suited for more homogeneous restrictions of the state-space. Following~\cite{DellaSantina2019DynamicObstacles} it is then possible to include further geometry and actuation based constraints, for a more aware control law.
        
        \begin{algorithm}[ht]
            \caption{Offline computation of state constraints.}\label{alg:constraint_finder}
            \begin{algorithmic}
                \State $q_l \gets [-\frac{\pi}{2}, -\frac{\pi}{2}, -\frac{\pi}{2}, -\frac{\pi}{2}]$
                \qquad $q_u \gets [\frac{\pi}{2}, \frac{\pi}{2}, \frac{\pi}{2}, \frac{\pi}{2}]$
                \If{check\_inclusion($q_l, q_u$)}
                    \State Use standard limits
                \Else
                    \State $q_l^* \gets [0, 0, 0, 0]$ \qquad $q_u^* \gets [0,0,0,0]$
                    \While{$i<N_{trials}$}
                        \State $q_l^t \gets $ Uniform\_Distribution($q_l, q_u$) 
                        \State $q_u^t \gets $ Uniform\_Distribution($q_l^t, q_u$)
                        \If{$(\Vert q_u^T - q_l^T\Vert_2>\Vert q_u^* - q_l^*\Vert_2)\quad \wedge$ \quad check\_inclusion($q_l^T, q_u^T$)}
                            \State New solution found
                            \State $q_l^* \gets q_u^T$ \qquad $q_u^* \gets q_l^T$
                        \EndIf
                        \State $i \gets i+1$
                    \EndWhile
                \EndIf
            \end{algorithmic}
        \end{algorithm}

\section{Simulation}\label{sim}

    \begin{figure*}[tb]
      \centering
      \includegraphics[trim = 100mm 20mm 100mm 20mm, clip, width = \linewidth]{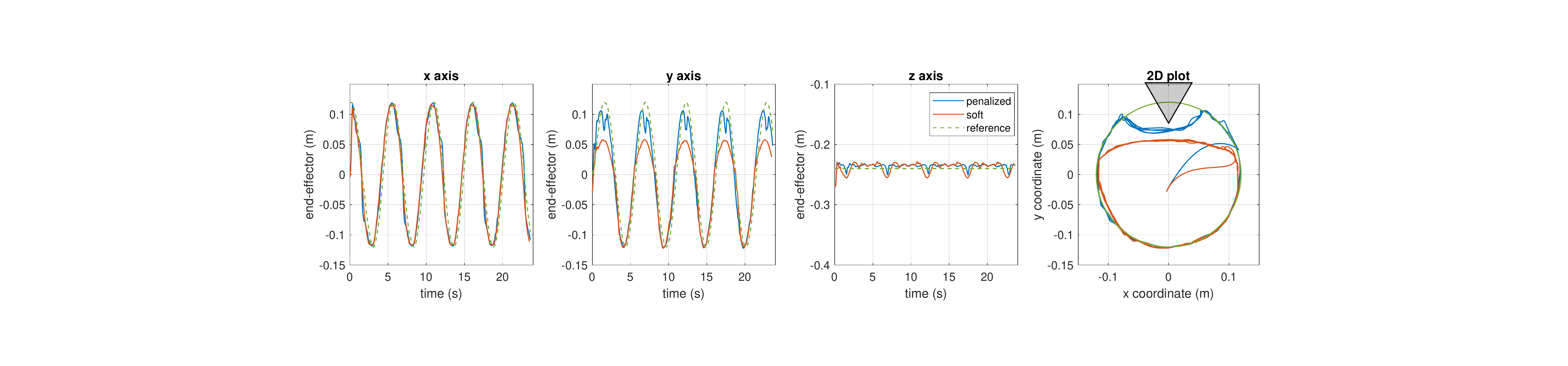}
      \caption{\emph{Robust MPC} simulation results. The penalized \emph{MPC} is shown in blue and the \emph{Soft-MPC} is shown in red. For sparse obstacles, the off-line computed set is more restrictive than the online penalization. The arm controller should follow the circular reference, but the presence of obstacles forces it to adapt.}
      \label{fig:1obstacle_sim}
      \vspace{-10pt}
    \end{figure*}
    
    \begin{figure}[tb]
      \centering
      \includegraphics[trim = 45mm 15mm 50mm 20mm, clip, width = \linewidth]{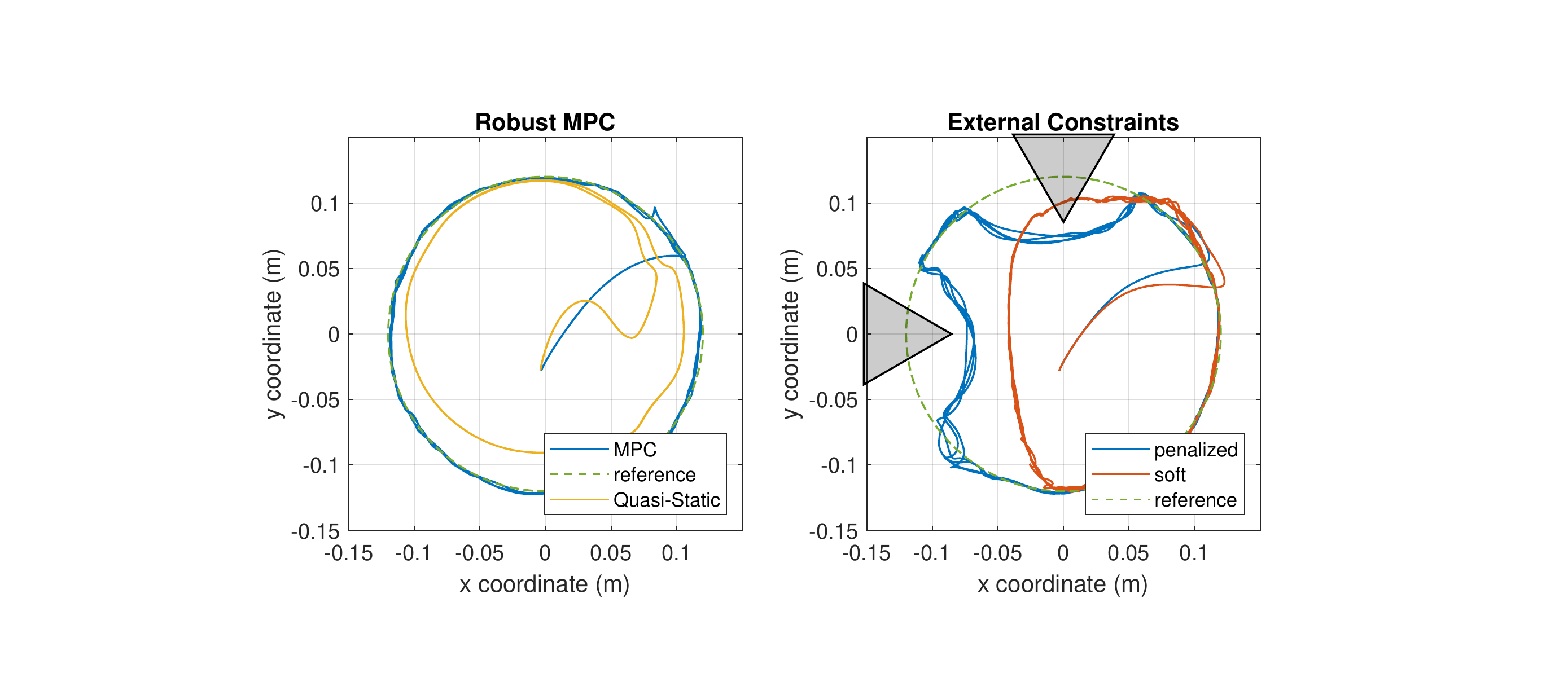}
      \caption{Simulation of circular trajectories. Left: given no additional constraints, tracking accuracy is high. A comparison with a quasi-static controller is provided. Right: a setting with sparse constraints is presented. The \emph{Soft-MPC} with offline computed set doesn't behave optimally under these conditions, since it is partially hitting the virtual obstacle due to its \emph{soft} implementation.}
      \label{fig:2obstacles_sim}
      \vspace{-10pt}
    \end{figure}
    
    We have run intensive simulations to validate our new method. The simulator is custom-made, working on a URDF model of the arm. Drake~\cite{drake} has been mainly used to solve the dynamics. As benchmark, we have considered the quasi-static controller, an approach similar to the one used in previous works for task-space references. It is based on inverse kinematics (no dynamic information) and small iterative updates to reduce the error; it is therefore slow and needs double the time of our MPC to follow the same trajectory, while running at the same rate. 
    
    In \cref{fig:1obstacle_sim} we have collected results of our \emph{Robust MPC} when performing a circular tracking while dealing with an obstacle at end-effector level. The penalized \emph{MPC} (option 1, in blue) performs ideally far from the obstacle and reacts to it only when within the neighborhood tuned in \cref{eq:penalized_MPC}. By adapting the penalization term, we can achieve larger or narrower turns around the obstacle, assumed symmetric. However, the same approach would work also for obstacles of more complex shapes, by penalizing all the corner positions. The \emph{Soft-MPC} of \cref{eq:soft_MPC}, red in the plots, instead delivers a more conservative output, where many more configurations are prevented from being reached. This difference is due to the nature of the offline problem we are solving, that isn't able neither to provide complex shapes nor to fully optimize the final set. These issues, that will be addressed in the future, are particularly evident for sparse obstacles configurations as the one analysed in \cref{fig:2obstacles_sim}. We can see how the \emph{Robust MPC} (on the left), that achieves an almost perfect tracking, can be adapted to account for obstacles, but the two options we provide behave differently. The penalized \emph{MPC} avoids areas of large cost, whose positions correspond to the obstacles and are defined either offline or online, while the \emph{Soft-MPC} has offline optimized over a joint variable set and works on it, even though could be extremely conservative for sparse obstacles configurations (as for the obstacle in position $(-0.14,0)$m), or possibly not effective (as happens for obstacle centered in $(0, 0.14)$m). In \cref{fig:2obstacles_sim} we can also observe how the quasi-static benchmark, despite using double the time, is still too slow to properly reach the required setpoint, since no dynamical information is available and therefore only almost static motions are possible.
    
    The \emph{Soft-MPC} with offline constraining has been designed to be applied in conditions of homogeneous restriction of the task space, for example a narrow hole along the length of the arm. This problem could represent a realistic setting where the robot needs to explore cavities and therefore being restricted in its motion capabilities. With our algorithm, joint configurations are limited to the only ones that allow enough motion at end-effector, while keeping almost constant the intermediate position. The results are shown in the video supporting this submission. Furthermore, using a more complex computation that features also mechanical information as in~\cite{DellaSantina2019DynamicObstacles}, resulting constraints would be even more appropriate.

\section{Physical Experiments}\label{exp}

    \begin{figure}[tb]
      \centering
      \includegraphics[trim = 20mm 20mm 40mm 20mm, clip, width = \linewidth]{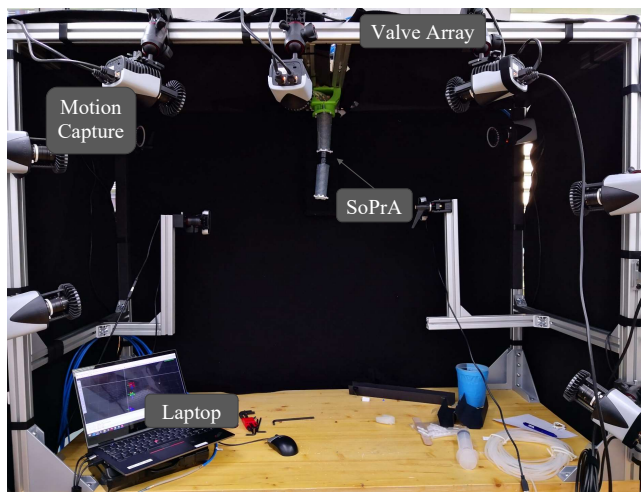}
      \caption{The SoPrA arm used consists of two fiber-reinforced segments made by three positive pressure chambers each. Between soft segments rigid connectors are present, and tubes are wired through them. A proportional valve array, placed above the arm, is used to independently pressurize the six cavities. The motion capture measures the segments' curvature.} %Adapted form~\cite{kazemipour2021robust}.
      \label{fig:setup}
      \vspace{-10pt}
    \end{figure}
    
    \begin{figure*}[tb]
      \centering
      \includegraphics[trim = 100mm 20mm 100mm 20mm, clip, width = \linewidth]{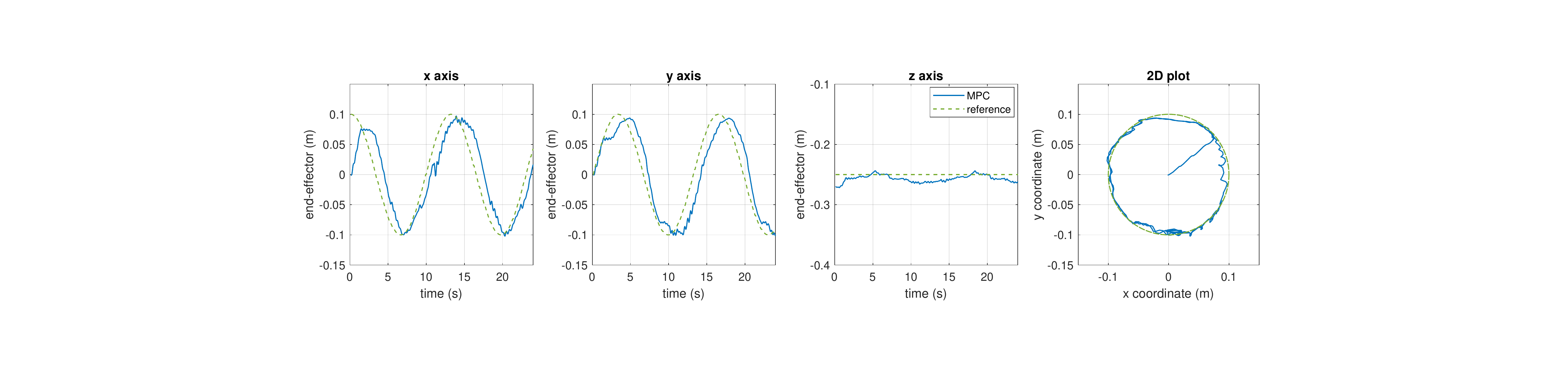}
      \caption{Real-world deployment of \emph{Robust MPC} on SoPrA. Controller performed two circular end-effector motions with a 10 cm radius. We observe a constant delay due to pressure input, but the tracking remains accurate. Controller runs at 15Hz.}
      \label{fig:robust_real}
      \vspace{-10pt}
    \end{figure*}
    
    \begin{figure}[tb]
      \centering
      \includegraphics[trim = 45mm 15mm 50mm 20mm, clip, width = \linewidth]{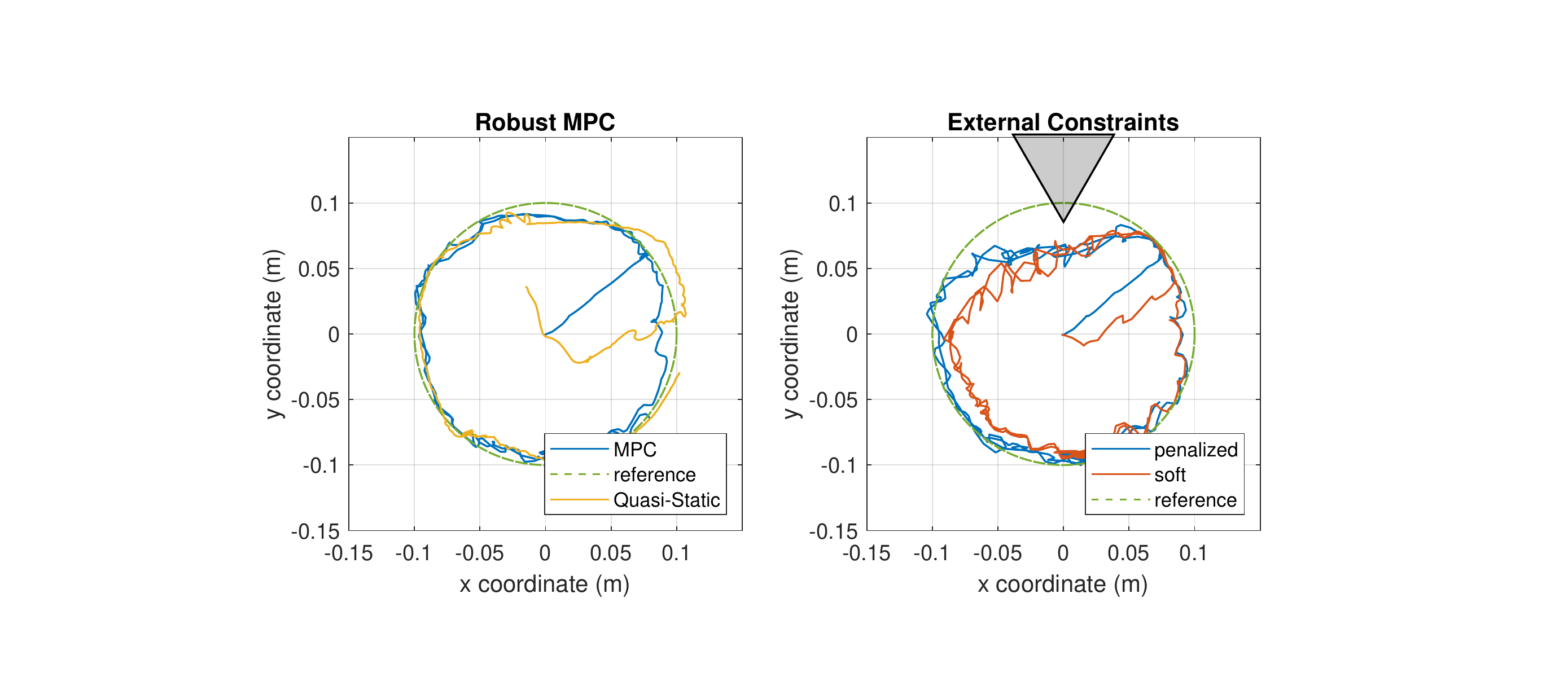}
      \caption{Real experiment showing two circular end-effector rotations in 20 seconds. Left: \emph{Robust MPC} without additional constraints. Right: penalized \emph{MPC} and \emph{Soft-MPC} with constrained set. Given this sparse obstacle setting, \emph{Soft-MPC} performs worse, but still achieves its goal accounting for external constraints. A benchmark for left plot is provided.}
      \label{fig:obstacles_real}
      \vspace{-10pt}
    \end{figure}
    
    \begin{figure*}[tb]
      \centering
      \includegraphics[trim = 100mm 20mm 100mm 20mm, clip, width = \linewidth]{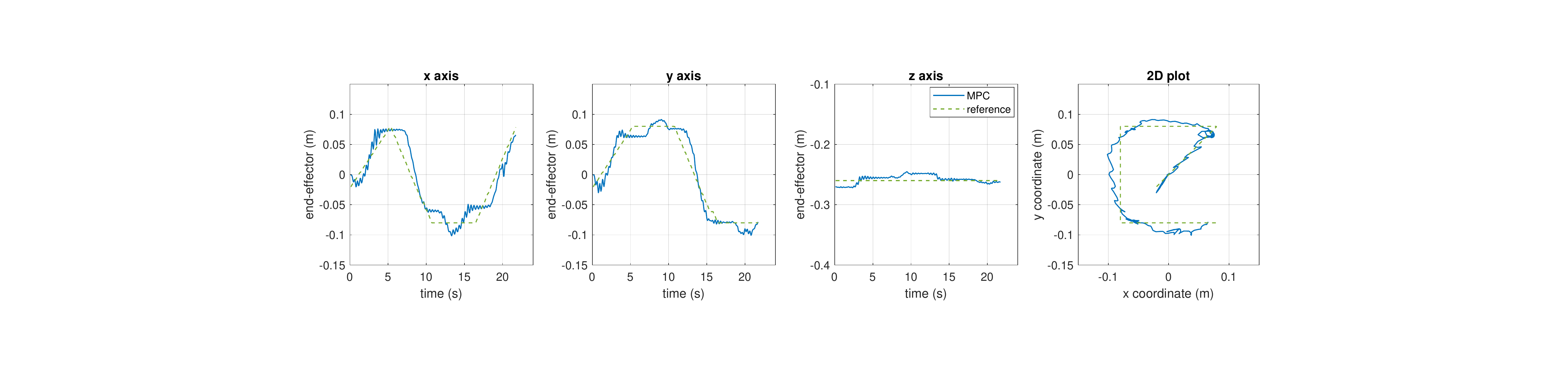}
      \caption{Square trajectory at constant height. SoPrA cannot reach the commanded positions, but \emph{MPC} can deal with the limits trying to deliver a trajectory that minimizes the error. Since the future trajectory is known, no corner is overshot.}
      \label{fig:square}
      \vspace{-10pt}
    \end{figure*}

    We have tested our control algorithm on a real robotic arm, SoPrA~\cite{Toshimitsu2021SoPrA:Sensing}. Instead of pursuing the best performance, we preferred to focus on showing the effectiveness of the concepts we have validated in simulation in a real environment, where model approximations and simplifying assumptions could have affected the control.
    In \cref{fig:setup} the experimental set-up has been summarized. Important to notice that the feedback is provided via a motion capture system made by infrared and color cameras. Reflective markers are attached in three positions along the arm: the base, the middle connector and the end-effector. The commanded pressures are fed to a proportional valve array that actuates the robot. 
    
    Our controller runs at \SI{15}{\Hz}, due to the limitations of the hardware used, with horizon length of 7 steps (the \emph{MPC} therefore predicts 0.5 seconds in the future). Control rate could be increased to almost 30Hz by implementing multi-threading and possibly further by selecting more efficient solvers. Throughout our experiments we have seen a consistent performance improvement with frequency increasing, so we believe that reaching a 50Hz control rate will allow this implementation to match other controllers' tracking accuracy. Although the low rate makes our controller prone to instability and with a low noise-rejection ability, we observe promising results thanks to the potential of the \emph{MPC} formulation in conditions where most other controllers would have failed. This implementation can therefore be applied in a vast range of situations where only cheap sensors are available, since it can work without the need of a high feedback rate. 
    
    The optimization variables $q, \dot{q}$ are initialized for the whole horizon with the measured value each iteration, while $u$ is initialized at zero. All dynamic parameters for \cref{eq:MPC_dyn} are numerically computed and provided to the solver each control loop. 
    
    In \cref{fig:robust_real} are reported the ideal results of our controller performing two turns of 10cm radius in 25 seconds. We observe a delay due to the pressure actuation that hasn't been addressed during the modeling phase. However, our implementation would allow to consider it with an approach similar to \cref{eq:4states_dyn}, where pressure dynamics is directly modeled. The future reference knowledge of \emph{MPC} would allow to account for the actuation delay and properly follow the trajectory variations, as no-other controller can do. On the 2D plot we observe a good tracking performance, with the controller able to damp out most of oscillations that occur due to the model mismatches and mechanical inhomogeneity.  
    
    In \cref{fig:obstacles_real} we show a more challenging setting, with 10 seconds turns in presence of obstacles. Both our proposed approaches, even thought affecting ideal performances, are able to deal with the more complex environment as expected from the simulation. We can however observe one drawback of the offline set computation, that prevents the arm from reaching the required position also where no obstacles are present; furthermore, we also observe a more oscillating behavior. These problems are due to the approach we chose in computing the set, that can constrain excessively the curvature variables. Checking the ideal performance only, the MPC formulation is comparable to quasi-static in terms of noise rejection, but allows for double the speed in reference tracking. Furthermore, MPC formulation can be used for additional tasks as shown in this paper.
    
    One interesting experiment is provided in \cref{fig:square}. Here, the commanded reference is a square lying outside of the reachability of SoPrA, that has a spherical task-space. Despite the inaccurate tracking performances, these results testify a few advantages that \emph{MPC} has over other controllers, extremely important to achieve realistic applications. First, our optimal controller is able to follow an unreachable reference just minimizing the error, without the need of reducing it to 0 (something that would happen with an integral term). Secondly, corners (abrupt reference variations) are never missed due to the future knowledge. This result is important for possible applications since the controller is more aware about its surroundings and can act accordingly. Lastly, even in presence of quite large model uncertainties, our robust formulation is able to keep the arm almost steady in position control, possibly with a bias. This result is expected from \emph{MPC} theory, and many ways to reduce the issue exist.  
    
    According to our simulated and physical results, we can certify the effectiveness of our approach, that allow us to include internal constraints for actuation limits and pressure dynamics, external constraints from the environment and admissible configurations dependent on shape and manufacturing in the same modular framework that makes easy to enable new features according to the task requirements. 
    We furthermore believe that \emph{MPC} would also allow to control underactuated robots with good performances; we achieved preliminary results towards this direction but we couldn't test intensively due to hardware and software limitations.

\section{Conclusion \& Future Work}\label{conclusion}
    We have shown how \emph{Model Predictive Control} can be adapted to soft robotics task-space control. \emph{MPC} is proposed as a way to handle many of the challenges typical of this control domain, such as actuation limits and reachable task-space limitations. We have built a framework based on the \emph{Augmented Rigid Body Model}~\cite{Katzschmann2019DynamicModel, della2020model} that can be used to control soft robotic arms also in prohibitive settings of low control rate and large model deviations. Our unified and modular approach allows to address many different issues at the same time, makes extremely easy to add new features and can be directly applied to arms of different shapes and actuation as long as an efficient kinematic representation can be retrieved. 
    
    This work represents a step towards new achievements in soft robotic arms control for practical applications, since controllers based on our method would allow the use of these robots in a realistic and cooperative environment, where safety and constraints' meeting are of primary importance. 
    
    Remaining issues to be addressed in future work are related to the large computation time and the model approximations; these have to be handled in order to increase the robustness in more challenging environments. To speed-up the control process many technical improvements can be employed, such as multi-threading or more advanced solvers. To face model approximations, our dynamics can be enriched with a data-driven oracle since the underlying \emph{MPC} framework makes it easy to integrate recently proposed learning methods as done in~\cite{Gillespie2018LearningNetworks,hyatt2020model}. The proposed framework would allow any future research to exponentially expand its applications, since better and more aware controllers might be designed, suitable for complex and potentially dangerous tasks that now are domain of rigid robotics only.
    Another long term goal is to increase the planning horizon of our \emph{MPC}; with this improvement our controller will be able to command faster motions by taking advantage of the stored potential energy when bending the silicon elastomer of soft robots, a unique capability of MPC framework.
    
    % In the future this controller could be applied to a model made with more PCC sections exploiting the MPC capabilities of dealing with underactuated systems. Also, MPC would allow to command faster motions able to take advantage of the potential elastic energy produced by silicon bending, a unique characteristic of soft robots. 
    % Our modular framework is so that learning methods can be applied with minimal effort. This would allow any future research to exponentially expand its applications, since better and more aware controllers might be designed, suitable for complex and potentially dangerous tasks that now are domain of rigid robotics only.
    
\section*{Acknowledgment}
    We thank Yasunori Toshimitsu and Oliver Fischer for their support, particularly their C++ codebase used for previous control approaches upon which we built this \emph{MPC} control framework. We also thank Amirhossein Kazemipour for his important insights during the planning phase of this work.

\bibliographystyle{IEEEtran}
\bibliography{references}

\end{document}